# Wood Species Recognition Based on SIFT Keypoint Histogram

Shuaiqi Hu, Ke Li
School of Information Science and Engineering
Southeast University
Nanjing, China

Xudong Bao
Laboratory of Image Science and Technology
Southeast University
Nanjing, China

*Abstract*—Traditionally, only experts who are equipped with professional knowledge and rich experience are able to recognize different species of wood. Applying image processing techniques for wood species recognition can not only reduce the expense to train qualified identifiers, but also increase the recognition accuracy. In this paper, a wood species recognition technique base on Scale Invariant Feature Transformation (SIFT) keypoint histogram is proposed. We use first the SIFT algorithm to extract keypoints from wood cross section images, and then k-means and k-means++ algorithms are used for clustering. Using the clustering results, an SIFT keypoints histogram is calculated for each wood image. Furthermore, several classification models, including Artificial Neural Networks (ANN), Support Vector Machine (SVM) and K-Nearest Neighbor (KNN) are used to verify the performance of the method. Finally, through comparing with other prevalent wood recognition methods such as GLCM and LBP, results show that our scheme achieves higher accuracy.

## I. INTRODUCTION

Wood, as a complex biological material, is used in a variety of fields. However, since different wood species have different proprieties, the usage of them varies a lot. For example, one species such as pine is suitable for house construction because it is strong and anticorrosive, but another species such as cottonwood is good for papermaking for its cheap price and abundance with wood fiber [1]. Correctly recognizing wood species definitely help people maximize the usage of wood. In addition, many precious wood has been exported illegally through mixing them with common wood species, which are similar in terms of shape and wood grain. Due to the difficulty to distinguish different wood species and the long time required to invite experts to recognize every patch of wood, wood smuggling has caused million dollars loss per year for countries such as Myanmar and Vietnam, and accelerated the extinction of several wood species [2]. Thereby, devising a wood species recognition technique that is portable and highly accurate is of great importance for timber industry.

Recently, wood recognition based on macroscopic images captured from wood cross section has been attached great importance for its simplicity and operability. Generally, these approaches to wood recognition can be divided into two types: focusing on different algorithm to extract texture features from wood images [3], [4], [5], or devising an effective method for feature selection and classification [6], [7], [8].

Several texture feature extraction algorithms have been adopted: M.Nasirzadeh applies Local Binary Pattern (LBP) algorithm to describe the texture feature of wood cross section images [4]. Wang constructs four gray level co-occurrence matrixes in different directions, and extracts six features: energy, entropy, inverse difference moment, dissimilarity, contrast, and variance from each matrix as texture descriptors [5].

Though wood recognition techniques has made unprecedented development, the recognition accuracy in practice is far from being satisfactory. The major cause lies in that these texture feature extraction algorithms compute features based on the whole image rather than the most distinguishable parts such as the distribution of wood grain and the array of pores. Especially, considering practical issues such as camera position change, illumination change and wood defects, extracting features from the whole image would destroy the intrinsic texture correlation and bring about irrelevance and redundancy.

In this paper, a new approach to wood species recognition based on SIFT keypoint histogram is proposed. Through the selection of keypoints, we use these scale invariant keypoints to represent the whole image to avoiding interference from irrelevant parts of the image. Then k-means algorithm is used to cluster the keypoints and k-means++ algorithm is adopted for initial seeds selection. After the clustering of keypoints, a statistical histogram of keypoints is calculated for each image as the final representation of the image.

The rest of this paper is organized as follows. Section II describes the method to represent an image with a SIFT histogram and the process of forming the histogram. Section III introduces several aspects of our experiment. Section IV shows the experimental results as well as necessary discussions. The paper ends in section V with some conclusions.

## II. SIFT KEYPOINTS HISTOGRAM

The procedure to compute SIFT keypoints histogram consists of three steps: SIFT keypoints extraction, Bag-of-Words method, and histogram calculation. The following parts of this section introduce these steps respectively.

### A. SIFT keypoints extraction

SIFT is an algorithm in computer vision for local feature detection and description in images. SIFT algorithm was proposed by David Lowe in 1999 [9], and further developed in 2004 by the same author [10]. Lowe's method describes an image with a large collection of feature vectors, each of which is robust to local geometric distortion and invariant to image rotation, translation and scaling. SIFT keypoints extraction consists of 2 steps: keypoints detection and keypoints description.

*1) Keypoints detection*

The first step of SIFT keypoints extraction is to detect keypoints of an image. Firstly, Gaussian filters at variant scales is convolved with the target image. Secondly, difference of successive Gaussian-blurred images are computed. Finally, Through comparing their numerical values, the maxima or minima points of the Difference of Gaussians (DoG) that appear at multiple scales are selected as keypoints, as shown in Fig. 1.

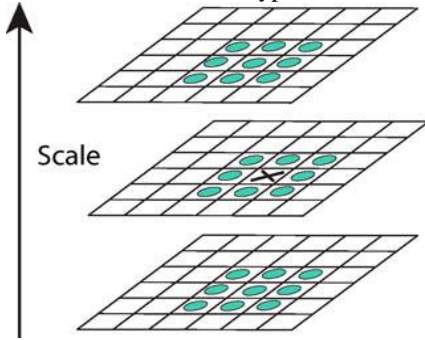

Figure 1. Through comparing one pixel (marked with cross) with its 26 neighboring pixels in 3×3 regions at the current and nearby scales (marked with circles), the maxima and minima of the difference-of-Gaussian images are detected.

*2) Keypoints description*

The second step is to describe the keypoint with 128 spatial orientation bins. The description of a keypoint is derived through several steps. Firstly, in the region that surrounds the keypoint, the orientation and gradient magnitude at each sample point of the image are computed, as illustrated in Fig. 2 (a), which are weighted by a Gaussian-based window and denoted by an overlaid circle. Secondly, the computation results are accumulated in orientation histograms which sum up the contents through $4 \times 4$ subregions, as illustrated in Fig. 2(b), and the length of every arrow indicates the sum of the gradient magnitudes within the region for that direction. Since we only care about the statistical meaning of keypoints, the position, orientation, and scale of the keypoint are not selected as descriptors in our scheme.

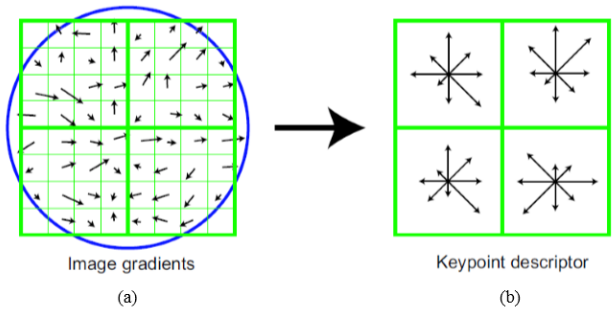

Figure 2. SIFT keypoints descriptor creation

*B. Bag-of-Words method*

The vocabulary of the Bag-of-Words method is learned from 128-dimension SIFT features of each keypoints from the training images. Clustering algorithm "k-means" is used to generate the vocabulary. This algorithm is also referred as Lloyd's algorithm for it was first proposed by Stuart Lloyd [11].

K-means algorithm aims for partitioning *n* observations into *k* classes in a way that every observation is classified to the class with nearest mean, which serves as a primitive prototype of the class.

For the first set of *k* means: $m_1^{[1]},…,m_k^{[1]}$, the algorithm is proceeded by carrying on the following two steps alternately [12]:

**Assignment step**: Assign every observation to the cluster that yields the smallest within-cluster sum of squares (WCSS) [13]. (From Mathematical concept, this means the observations are partitioned according to the Voronoi diagram [14], which is produced by the means).

$$S_i^{[t]} = \left\{x_p : \| x_p - m_i^{[t]} \|^2 \leq \| x_p - m_j^{[t]} \|^2 \; \forall j, 1 \leq j \leq k\right\},$$

**Update step**: Update the centroids of the observations in the new clusters with the calculated new means.

$$m_i^{(t+1)} = \frac{1}{|S_i^{(t)}|} \sum_{x_j \in S_i^{(t)}} x_j$$

Since k-means algorithm might lead to convergence to local minimum and thus produce counterintuitive results, k-means++ algorithm, proposed by David Arthur and Sergei Vassilvitskii in 2007 [12], is adopted for selecting the initial values (or "seeds") for k-means clustering algorithm. The detailed algorithm is presented as follows:

1) Among the data points, select one center uniformly at random.
2) For every data point *x*, calculate its distance to the nearest already chosen center and denote this distance as D(*x*).
3) Randomly select a new data point as the new center. The selection criteria is based on weighted probability distribution, in which the possibility of a certain point *x* to be selected is proportional to $D(x)^2$
4) Carry on Step 2 and 3 alternately until *k* centers are selected.
5) Since the initial centers have been selected, k-means clustering can be adopted with little possibility of convergence to local minimum.

*C. Histogram counting*

For each image, its keypoints are classified into *k* clusters. Then a histogram that counts the occurrence of each cluster is computed for every image. Since different images are detected with different amount of keypoints, the occurrence of each cluster divides the total number of keypoints to serve as the probability of occurrence of each cluster and these probabilities form the final SIFT keypoint histogram for each image.

## III. EXPERIMENT ASPECTS

### A. Wood Dataset

The wood samples in our experiment consist of macroscopic images from common wood species in timber industry as well as some precious wood species such as Myanmar gold camphor, Diospyros crassiflora, etc. The image acquisition process is implemented using SONY ILCE-6000 camera with Canon MP-E 65 mm macro lens. Each image is acquired with 22 aperture and 30mm focus. There are 28 species of wood in this dataset with 100 images in each. The image size is $3000 \times 2000$ pixels and we resize each image to $600 \times 400$ pixels for calculation reduction. Some samples of the dataset is shown in Fig. 3.

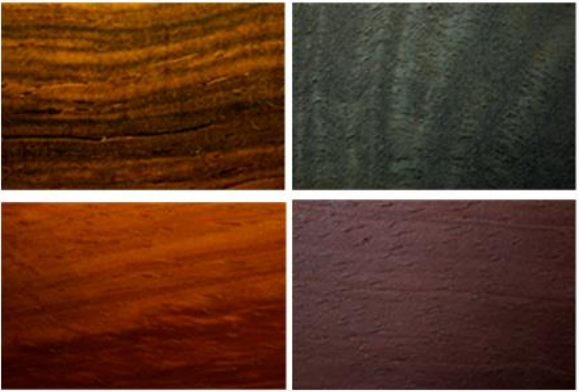

Figure 3. Samples of macroscopic wood images in the dataset

### B. Classification Model

Three kinds of classification models are used as classifiers in our experiment: Artificial Neural Networks (ANN), Support Vector Machine (SVM) and k-Nearest Neighbor (K-NN). ANN has been successfully applied in solving various classification tasks such as authentication of coins [15], sign language recognition [16], and EKG pattern recognition [17]. For high dimensional data set, SVM has been proven to be a pretty effective classification method, especially in face recognition [18], and object recognition [19]. In addition, some wood species recognition systems adopted k-NN as classifiers [20], [4]. For ANN, in each species, 60 images are used for classifier training, 20 images are used for validation and the remaining 20 images are used for testing. For SVM and k-NN, in each species, 80 images are used for classifier training and the remaining 20 images are used for testing the trained classifier. Besides, cross validation, which selects a number of folds to partition the data into, is chosen as the validation method.

#### 1) Artificial Neural Networks (ANN)

Inspired by animal brains' neural structure, artificial neural networks, referring to relatively crude electronic networks of "neurons", process records one at a time and "learn" by comparing their classification results with recorded actual classification results. The error from the initial classification of the first record is fed back into the ANN, and used in a specific way to modify the network algorithm for second iteration, and so on for many iterations [21].

Backpropagation learning algorithm is applied to obtain the weight of the network and *trainscg*, which updates bias values and weight according to the scaled conjugate gradient method, is selected as the backpropagation function. To avoid the interference of hidden layer number and maximize the performance of ANN, ANN with hidden layers set to 60, 80, 100, 120, and 140 are implemented.

#### 2) Support Vector Machine (SVM)

Seeking a hyperplane that maximizes the margin between two classes, SVM uses kernel functions to project data to a higher dimensional space for more separable data [22]. In our experiment, five kinds of kernel functions are used. These kernel functions are linear function, medium Gaussian function, coarse Gaussian function, quadratic function, and cubic function.

#### 3) K-Nearest Neighbor (K-NN)

K-NN compares the features of the test sample against all the training samples and chooses k samples with the shortest distance [23]. In our experiment, three kinds of distinctions between classes are used: fine k-NN (the number of neighbors set to 1), medium k-NN (the number of neighbors set to 10) and coarse k-NN (the number of neighbors set to 100) with Euclidean distance. In addition, k-NN with weighted distance is also implemented.

### C. Processing Steps

Since Lowe's SIFT scheme is designed for gray scale images, the first step is to convert the RGB images to gray scale images. Then the keypoints as well as their descriptors of each image are extracted using SIFT algorithm. Furthermore, the descriptors of all the keypoints from images in the training set are piled up as raw data, which are $28 \times 80 = 2,240$ images. These descriptors are clustered using k-means algorithm into k clusters and k-means++ algorithm is used for the selection of initial seeds. Then each image in the training set calculates the probability of the occurrences of each cluster in its keypoints and forms an SIFT keypoint histogram. Finally, several classifiers, including ANN, k-NN and SVM, are used for classification. For testing images, the process steps are similar to that of the training images. However, in order to avoid interference, the keypoints of the test images are not involved in the k-means clustering process and these keypoints are clustered using the results of the keypoints clustering from training images. A block diagram illustrating the process of our proposed method is shown in Fig. 4.

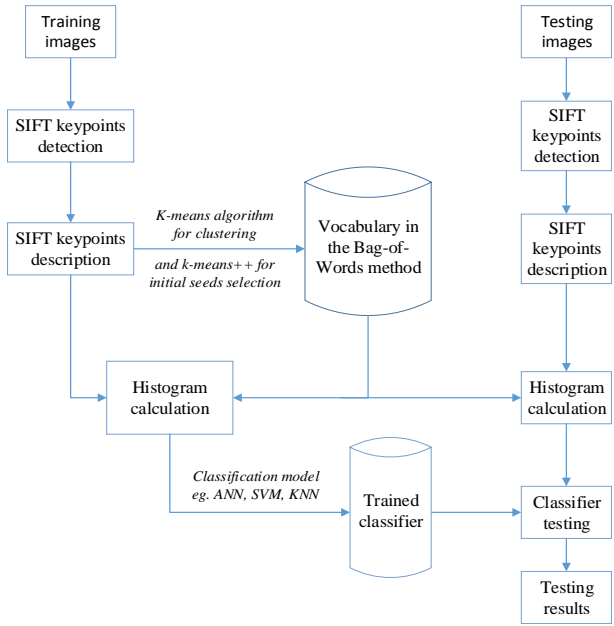

Figure 4. The procedures of the proposed wood recognition system

## IV. EXPERIMENTAL RESULTS AND DISCUSSIONS

In this section, two experiments are discussed. Experiment 1 is conducted to test the influence of the amount of k-means clusters on the performance of our scheme. Experiment 2 explores the classification accuracy of other wood recognition schemes using our datasets and compares the performance of these schemes with ours.

### A. Experiment 1: Selecting the Best Class Numbers and Best Classifier for Better Accuracy Rate

In this experiment, 4 kinds of cluster quantity is implemented: 250 clusters, 300 clusters, 350 clusters, and 400 clusters, using different kinds of classificaiton methods. The experiment results are shown in Fig. 5.

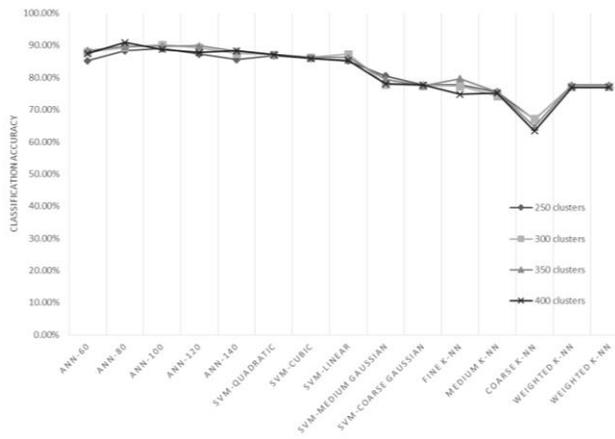

Figure 5. The performance of our scheme using different classifier and with different amount of clusters

Judging from the results of the experiement, our scheme performs best while using ANN or Quadratic kerneled SVM. In addition, when the cluster quantity is large enough for clusters to represent the distribution of keypoints, the difference in the amount of clusters do not have a significant infulence on the actual classification accuracy.

### B. Experiment 2: The Comparison Between SIFT keypoints histogram and other prevalent wood recognition schemes

The objective of this experiment is to compare our scheme with some prevalent wood recognition schemes. Two other wood texture extraction algorithm (LBP and GLCM) are implemented, which are introduced by Nasirzadeh [4] and Wang [5] respectively. In this experiment, both schemes are implemented in the way exact to the method introduced in [4] & [5], but the dataset is changed to our own dataset, which is introduce in part A, section III. The number of features for these two scheme as well as our scheme with 300 clusters is shown in table I.

TABLE I. NUMBER OF FEATURES FOR EACH METHOD

| Method | Number of Features |
|---|---|
| SIFT keypoints histogram | 300 |
| LBP | 256 |
| GLCM | 24 |

The classification accuracy of these schemes is illustrated in Fig. 6.

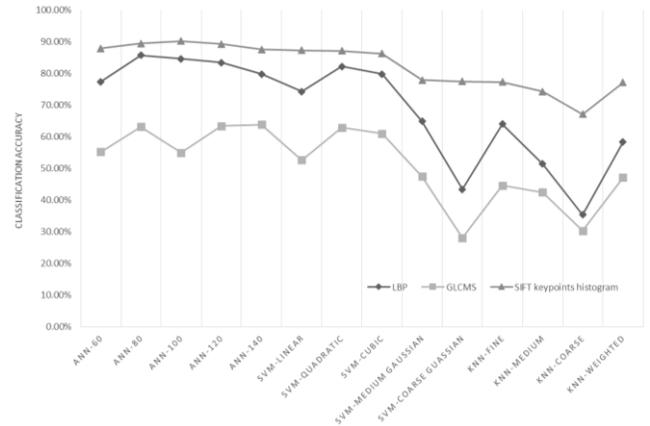

Figure 6. Classification accuracy of LBP, GLCM, and SIFT keypoints histograms

The best classification accuracy of these wood recognition schemes, usinng ANN, SVM and k-NN, is shown in table II.

TABLE II. CLASSITION ACCURACY OF DIFFERENT SCHMES

| Classification Accuracy | Wood Recognition Schemes | | |
|---|---|---|---|
| | SIFT keypoints histogram | LBP | GLCM |
| ANN | 90.20% | 85.72% | 63.81% |
| SVM | 87.50% | 82.20% | 62.90% |
| k-NN | 77.32% | 64.00% | 47.10% |

As shown above, with these recognition schemes, k-NN performs worst, SVM performs better, and ANN produces the most desirable results. Compared to algorithms that extract texture features from the whole images such as LBP and GLCMs, our scheme performs significantly better, no matter what classification method is used. In addition, the higher accuracy performed by our scheme might due to the fact that compared to LBP (256 features) and GLCM (24 features), the number of features in our scheme (300 features) is large enough to aid classifiers for better classification. Perhaps, most importantly, in our scheme, the number of features is adjustable rather than unmodifiable, which means when a larger dataset is adopted, the number of features can be adjusted larger to describe the texture features in more specific details and thus produces better classification accuracy.

## V. Conclusion

This paper has proposed a new method for wood species recognition based on SIFT keypoints histogram. In this method, texture features are extracted from distinguishable parts of wood images rather than from the whole image, thereby the interference of the camera position change, illumination change and wood defects are avoided. The experimental results demonstrate that this method achieves higher accuracy than LBP and GLCM. The correct ratio can reach up to 90.2%, which basically fulfills the requirement of practical wood recognition.

### Acknowledgment

This work is supported by National Undergraduate Training Program for Innovation and Entrepreneurship under grant 201510286017 and Laboratory of Image Science and Technology of Southeast University